\documentclass{article} 
\usepackage{styles/iclr2025_conference,times}

\usepackage{amsmath,amsfonts,bm}

\def\eqref#1{equation~\ref{#1}}

\def\1{\bm{1}}

\DeclareMathAlphabet{\mathsfit}{\encodingdefault}{\sfdefault}{m}{sl}
\SetMathAlphabet{\mathsfit}{bold}{\encodingdefault}{\sfdefault}{bx}{n}

\usepackage[utf8]{inputenc} 
\usepackage[T1]{fontenc}    
\usepackage{hyperref}       
\usepackage{url}            
\usepackage{amsfonts}       
\usepackage{nicefrac}       
\usepackage{microtype}      
\usepackage{xcolor}         
\usepackage{booktabs} 
\usepackage{multirow}
\usepackage{graphicx} 
\usepackage{amsmath}
\usepackage{xspace}
\usepackage{enumitem}
\usepackage[capitalize]{cleveref}

\newcommand{\model}{Dragonfly}
\newcommand{\modelbiomed}{Dragonfly-Med}

\title{{\model}: Multi-Resolution Zoom-In Encoding Enhances Vision-Language Models}

\author{%
  \textbf{Rahul Thapa$^{1,2}$}\thanks{Equal contribution. Corresponding author: \texttt{rthapa84@stanford.edu}} \quad \textbf{Kezhen Chen$^{1}$}\footnotemark[1] \quad \textbf{Ian Covert$^{2}$} \quad \textbf{Rahul Chalamala$^{1,3}$} \quad \textbf{Ben Athiwaratkun$^{1}$} \\
  \textbf{Shuaiwen Leon Song$^{1}$} \quad \textbf{James Zou$^{1,2}$} \\
  $^1$Together AI \quad $^2$Stanford University \quad $^3$Caltech \\
}

\iclrfinalcopy 
\begin{document}

\maketitle

\begin{abstract}

Recent advances in vision-language models (VLMs) have demonstrated the advantages of processing images at higher resolutions and utilizing multi-crop features to preserve native resolution details. However, despite these improvements, existing vision transformers (ViTs) still struggle to capture fine-grained details from less prominent objects, charts, and embedded text, limiting their effectiveness in certain tasks. In this paper, we extend recent high-resolution and multi-crop techniques by not only preserving the native resolution, but zooming in beyond it and extracting features from a large number of image sub-crops. This enhancement allows our model to better capture fine-grained details, overcoming the limitations of current ViTs. To manage the increased token count and computational complexity, we demonstrate that a simple mean-pooling aggregation over tokens is effective. Our model, {\model}, achieves competitive performance on general-domain tasks such as ScienceQA and AI2D, and excels in tasks requiring fine-grained image understanding, including TextVQA and ChartQA. Among models in the 7-8B parameter range, {\model} consistently ranks at the top across ten general-domain benchmarks, achieving the highest or second-highest scores in most cases, outperforming models that are significantly larger or trained on larger datasets. Our biomedical model, {\modelbiomed}, sets new benchmarks on several medical tasks, achieving 91.6\% accuracy on SLAKE (compared to 84.8\% for Med-Gemini), a 67.1\% token F1 score on Path-VQA (compared to 62.7\% for Med-PaLM M), and state-of-the-art results across the majority of image captioning tasks. Overall, our work highlights the persistent challenge of engineering visual representations with fixed-resolution ViTs, and proposes a simple yet effective solution to address this issue and boost performance in both general and specialized domains. Our codebase is available at \url{https://github.com/togethercomputer/Dragonfly}.

\end{abstract}

%
%

\section{Introduction}
\label{section_introduction}

Vision-language models (VLMs) represent an exciting and rapidly evolving field, offering new possibilities for open-ended visual tasks by leveraging the knowledge and reasoning abilities of large language models (LLMs). Ongoing research explores how best to integrate visual information into LLMs, with many recent advances using image encoders to map visual data into the latent space of LLMs by dividing images into patch-level tokens, which are then aligned with the LLM during visual instruction tuning \citep{llava, llava-next, yang2023mm, li2023otter, llava-uhd, mm1, idefics2, ferret, ferretv2}.

Early VLMs processed images at fixed, low resolutions, requiring high-resolution images to be downsampled to fit model input dimensions \citep{llava, alayrac2022flamingo}. This downsampling often causes shape distortion, loss of fine details, and reduced overall visual richness—especially for tasks that demand fine-grained visual understanding. To address this limitation, recent works have demonstrated the benefits of using higher-resolution encoders, where avoiding excessive downsampling improves performance across various tasks \citep{qwenvl, ferretv2, chen2023pali, idefics2, mm1}. Moreover, approaches such as Llava-1.5 \citep{llava-next} and Llava-UHD \citep{llava-uhd} use multi-crop techniques, where an image is divided into multiple crops, enabling models to process images at or near their native resolution. This aligns with the conventional wisdom in computer vision that preserving images near their original resolution retains crucial information, which is vital for tasks requiring fine-grained visual understanding, such as text recognition in charts or other dense visual content \cite{li2024monkey, beyer2024paligemma, mm1}.

In this paper, we extend the high-resolution encoding approach by introducing a novel strategy: featurizing images with multiple crops that zoom beyond the native resolution. By magnifying images to this level, we aim to address the limitations of existing vision transformers (ViTs), particularly their difficulty in extracting fine-grained details from less prominent objects, charts, and embedded text \citep{li2023otterhd, qwenvl, hong2024cogagent, ye2023ureader}. While one might expect that zooming beyond native resolution adds no additional information and should not help if ViTs are functioning perfectly, in practice, they often miss subtle image details. As a result, zooming in helps capture information that ViTs currently struggle to extract. However, this high-resolution zoom-in and multi-crop method presents a new challenge: the number of image tokens increases with higher resolutions and additional crops, significantly expanding context length and computational demands. For example, images are converted into 576 visual tokens using the CLIP ViT-L/14@336px backbone \citep{radford2021learning}. With five image crops, this number already exceeds 2,800 tokens \citep{llava-next}, and our zooming in beyond native resolution requires substantially more crops. To manage this token complexity, we explore various options to compress visual tokens into a manageable context length. Empirically, we find that a simple mean-pooling approach is effective, and we adopt this method in our model.

In summary, \textbf{our contributions} are as follows:

\begin{itemize}[leftmargin=1.5em]

    \item We introduce {\model}, a new large VLM that processes images using multiple image crops that zoom beyond native resolution. By employing simple mean-pooling aggregation on high-resolution crops, {\model} efficiently reduces visual token counts while preserving fine-grained image details. {\model} excels on general-domain benchmarks such as ScienceQA and AI2D, and performs especially well in tasks requiring fine-grained image understanding, like ChartQA and TextVQA. Among models in the 7-8B parameter range, {\model} consistently ranks at the top across ten general-domain benchmarks, achieving the highest or second-highest scores in most cases, outperforming models that are significantly larger or trained on larger datasets.

    \item We highlight the model's strong performance on biomedical tasks, where detailed image comprehension is critical. When fine-tuned on a biomedical instruction-tuning dataset, {\modelbiomed} achieves state-of-the-art or competitive results across benchmarks such as VQA, image captioning, and radiology report generation. Notable outcomes include 91.6\% accuracy on SLAKE, a 67.1 token F1 score on Path-VQA, and a 50.9 CIDEr score on MIMIC-CXR captioning—these are the highest reported numbers to the best of our knowledge.
    
    \item We curate a dataset of 2 million supervised fine-tuning samples for the general domain and 1.4 million for the biomedical domain. While most of the data is publicly available, we carefully balance and deduplicate the dataset across multiple tasks. For the biomedical domain, we also ensure the dataset is balanced across various image modalities. Our codebase is available at \url{https://github.com/togethercomputer/Dragonfly}, and both models are released on HuggingFace.
    
\end{itemize}

\begin{figure}[t]
    \centering
    \includegraphics[width=\textwidth]{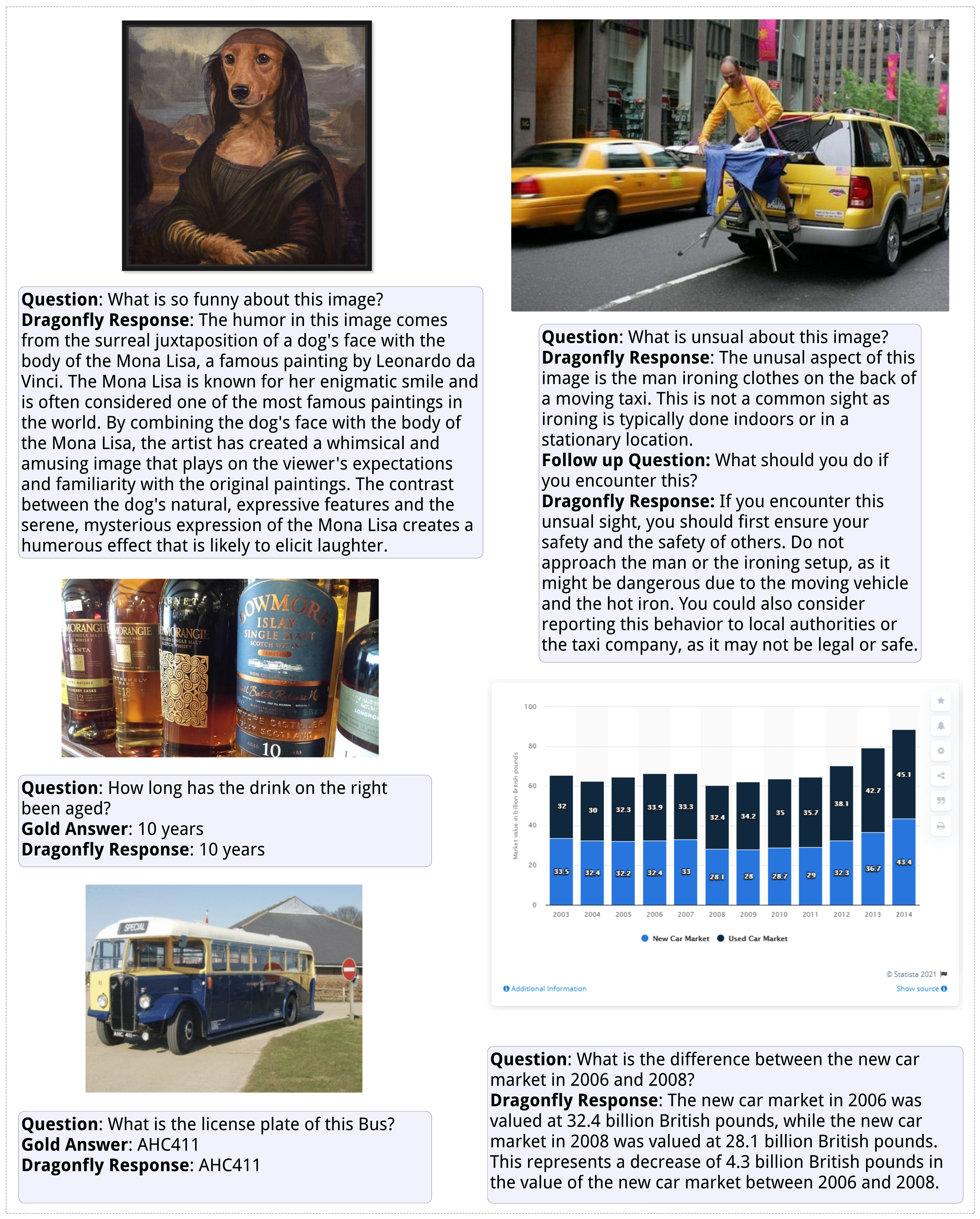}
    \caption{Examples generated by {\model}, showcasing its diverse capabilities, including world knowledge and humor, multi-turn question-answering, OCR, and chart understanding. }
    \label{fig:general_domain_examples}
\end{figure}

%
%

\section{Related Work}
\label{section_related_work}

\paragraph{Vision-Language Models (VLMs)} The advancement of VLMs has significantly impacted artificial intelligence, extending the reasoning capabilities of LLMs to the visual domain. Most of these models are developed through visual instruction-tuning, which merges vision and language by integrating pretrained ViTs and LLMs \citep{llava,llava-next,dai2023instructblip,yang2023mm,li2023otter,llava-uhd,mm1,idefics2,ferret,openflamingo}. For example, \citet{llava} employs a fully connected layer to map image embeddings produced by a pretrained CLIP encoder \citep{radford2021learning} into the embedding space of an LLM \citep{vicuna2023}. This straightforward approach has enabled the emergence of powerful capabilities, such as visual question answering, image captioning, and multimodal reasoning, allowing models to interpret and interact with both visual and textual data simultaneously. Important design choices within this approach are still being investigated, and many models still downscale input images to fixed, low resolutions, leading to the loss of fine visual details. 

\paragraph{High-Resolution Inputs and Fine-Grained Details} Handling high-resolution inputs in VLMs presents significant challenges, particularly due to the increase in image tokens, which leads to greater computational demands. Recently, a multi-crop approach has emerged that attempts to overcome the limitations of ViTs processing images at a fixed resolution \citep{llava}. However, using multiple crops significantly increases the number of visual tokens. LLaVA-UHD \citep{llava-uhd} is another multi-crop approach that segments native-resolution images into smaller slices to retain detailed visual information while managing the number of tokens using a perceiver resampler \citep{perceiver}. Additionally, approaches like Qwen-VL \citep{qwenvl}, PaLI-3 \citep{chen2023pali}, and PaLI-X \citep{chen2023pali} have been explored to gradually scale input resolution, but they must still downsize large images and potentially lose critical information. Additionally, capturing fine-grained, local details—essential for tasks like segmentation and object detection—remains a challenge for models like CLIP, which are trained on global image-level captions and often miss important regional semantics \citep{wu2023clipself, xu2022simple, zhong2022regionclip}. One potential way to overcome these limitations is to zoom in beyond the native resolution of an image, which enables models to extract even finer details that may not be captured at standard resolutions. By focusing on smaller regions of the image at higher magnification, this approach helps to compensate for the shortcomings of current ViTs in capturing localized features that may be lost when images are processed in a single fixed-resolution crop.

\paragraph{Biomedical Applications of VLMs} 
VLMs have shown significant potential in general domains, sparking growing interest in their application to various biomedical tasks. Models such as BiomedGPT \citep{zhang2023biomedgpt} and LLaVA-Med \citep{li2024llava} integrate medical imaging and text from the scientific literature to address specialized tasks in multiple biomedical domains. General-purpose models have also been adapted for medical applications, including Med-PaLM \citep{tu2024towards}, Med-Flamingo \citep{moor2023med} and Med-Gemini \citep{saab2024capabilities}, showcasing the potential of VLMs to tackle complex vision-language tasks.

%
%

\section{{\model} Architecture}
\label{section:method}

\begin{figure}[t]
    \centering
    \includegraphics[width=\textwidth]{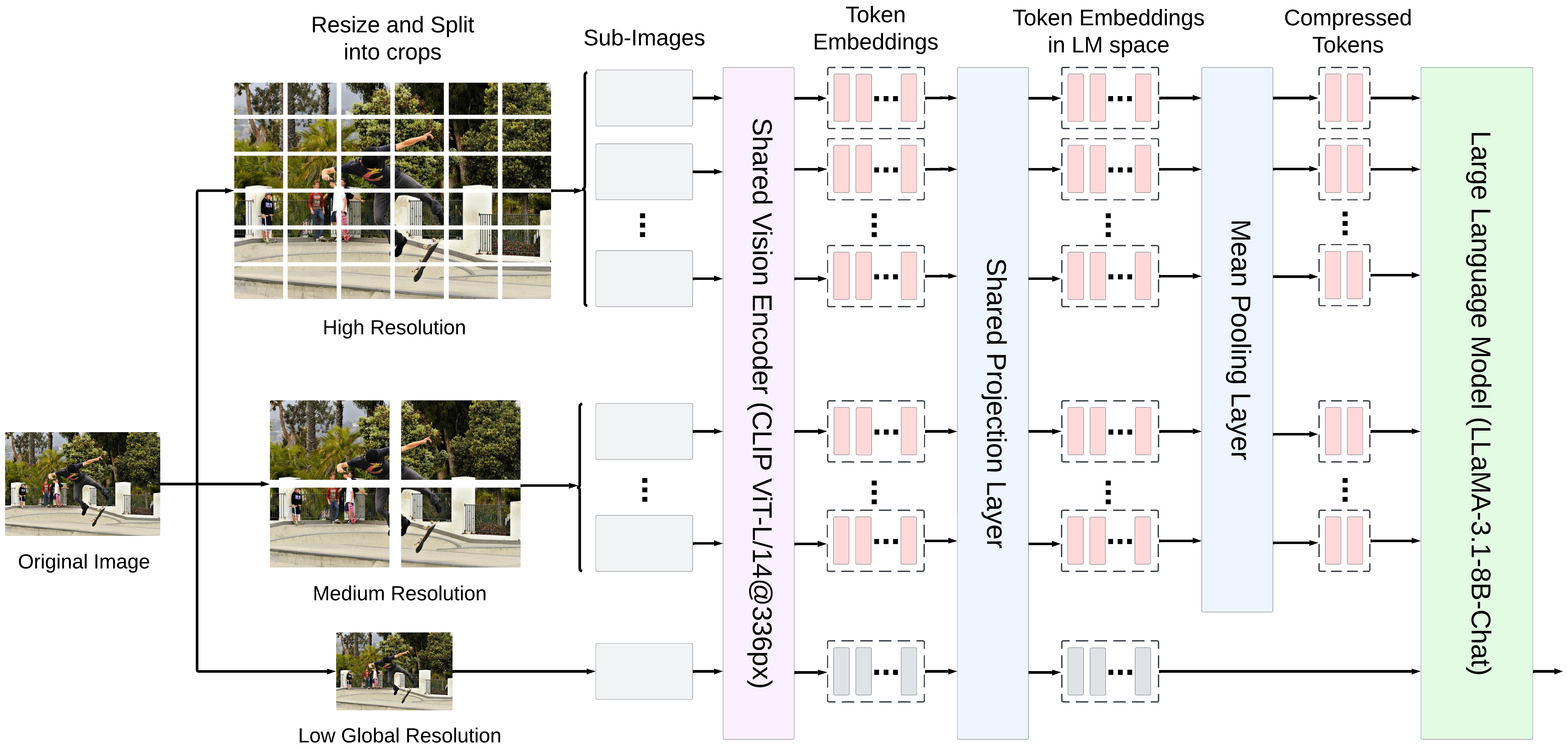}
    \caption{Overview of our proposed {\model} framework. The original image is resized into low, medium, and high resolutions. The medium- and high-resolution images are divided into crops matching the encoder's training resolution. All sub-crops pass through a shared vision encoder to produce visual tokens. The projection layer then maps the visual tokens to the language space. Afterward, the mean-pooling layer reduces the embeddings from each sub-crop into 36 tokens.}
    \label{fig:together_model_overview}
\end{figure}

We introduce our multi-crop visual encoding approach and the strategies employed to manage the large number of visual tokens resulting from it. The workflow of our architecture is illustrated in \cref{fig:together_model_overview}.

\subsection{Multi-resolution Visual Encoding}
\label{sub-sec:multi-resolution}

We employ a multi-resolution visual encoding strategy using a shared image encoder trained on a fixed resolution of $R \times R$. Following techniques from previous works \citep{llava-next, llava-uhd}, our framework processes larger images by dividing them into multiple crops, each matching the encoder's expected resolution. Specifically, given an image $I$, we resize it into three distinct resolutions: a low-resolution image $I^l$ of size $R \times R$, a medium-resolution image $I^m$ of size $x^{m}R \times y^{m}R$, and a high-resolution image $I^h$ of size $x^{h}R \times y^{h}R$. The medium- and high-resolution images are then divided into crops, resulting in two sets of crops, $\{I^m_i\}_{i=1}^{x^{m} \times y^{m}}$ and $\{I^h_j\}_{j=1}^{x^{h} \times y^{h}}$, with each sub-image aligned to the encoder's training resolution $R \times R$. We adopt a simplied version of any-resolution segmentation method from \citet{llava-uhd} to divide images into crops. This method selects a resolution grid from a pre-defined set of grids that approximately match the original image's aspect ratio. For medium resolution, the possible grids are $\{(2,2),(1,4),(4,1)\}$, which each result in four crops. For high resolution, we use the grids $\{(6,6), (3,12), (12,3)\}$, which each produce 36 crops in total.

The image encoder encodes each sub-image into a sequence of visual tokens $\{v_1, \dots, v_n\}$. These tokens, extracted from the various crops, are projected into the latent space of the language model via a projection layer $P$, generating a corresponding sequence of projected tokens $\{t_1, \dots, t_n\}$. The projected tokens from different crops are concatenated to form a comprehensive representation of the image, which is then used for understanding by the LLM. However, due to the large number of crops, especially from the high-resolution set, incorporating all these crops can result in longer context lengths that either exceed the capacity of many LLMs or become prohibitively expensive. In the following sections, we discuss strategies to mitigate these challenges.

\subsection{Token Aggregation}
\label{sub-sec:token-strategy}

Our objective is to compress visual information from medium- and high-resolution images while preserving the fine-grained details from these magnified views. This compression is essential to manage the increased number of tokens generated from multiple high-resolution crops, which would otherwise result in excessive computational demands. Based on our experiments, we find that a simple mean pooling strategy effectively reduces the number of visual tokens while still retaining the benefits of zooming to higher resolutions.

All images are resized to $336 \times 336$ and processed using the CLIP ViT-L/14@336px model, which outputs 576 tokens. For the low-resolution image, we retain all 576 tokens. For the medium- and high-resolution images, the image is divided into 40 total crops (4 for medium resolution and 36 for high resolution). Each sub-image is encoded with 576 tokens, which are reshaped into a $24 \times 24$ token grid. We then compress this representation, applying mean pooling with stride 4 to reduce it to a grid of size $6 \times 6$, or 36 tokens per sub-image. All 40 crops are then concatenated with separator tokens placed between them, forming the complete image representation that is passed to the LLM. This includes 576 tokens from the low-resolution image, $4 \times 36=144$ tokens from the medium resolution, and $36 \times 36=1,296$ tokens from the high resolution, yielding a total of 2,016 image tokens. 

%
%

\section{Experiments and Results}
\label{section:experiments}

In this section, we first introduce our experimental setup, and then present ablations and baseline comparisons to validate our design choices for multi-resolution visual encoding. Next, we evaluate {\model} against other models of similar scale across multiple general-domain benchmarks. Finally, we continue training {\model} on a biomedical dataset, resulting in {\modelbiomed}, and assess its performance on biomedical tasks.

\subsection{Experimental Setup}
\label{subsection:experimental_setup}

{\model} uses Llama3.1-8B-Chat \citep{llama3} as the language backbone and CLIP ViT-L/14@336px \citep{radford2021learning} as the image encoder. CLIP ViT-L/14@336px accepts images with a resolution of $336 \times 336$, and our highest resolution is either $2016 \times 2016$ or $1008 \times 4032$, depending on the aspect ratio of the image. An analysis of the resolutions across our training data revealed that with out multi-resolution encoding approach, 99.5\% of images are magnified beyond their native resolution, 95\% of the images are zoomed in by at least $2 \times$, and 65\% are zoomed in by at least $4\times$. A cumulative density plot of the zoom-in ratio for images in our dataset is provided in \cref{fig:max_res_kde}.

For training {\model}, we adopt the two-stage visual instruction-tuning framework introduced by \citet{llava}. In the first stage, the LLM and vision encoder are frozen, with only the projection layer being trained. This stage allows the projection layer to effectively learn how to map visual tokens into the language space. The model is trained for one epoch on the LLaVA-Pretrain dataset \citep{llava}, which consists of 558K image-text pairs, using a global batch size of 64 and a learning rate of 2e-5.

In the second stage, the entire model undergoes fine-tuning on a high-quality visual instruction-tuning dataset. During this stage, the LLM learns to process visual information, thereby optimizing the model's performance in vision-language tasks. For this supervised fine-tuning, we curated a dataset comprising 2M image-instruction samples from various sources, which include detailed image descriptions, complex reasoning tasks, and diverse visual question-answering tasks. Further details about the general domain instruction-tuning dataset are provided in \cref{app:section_general_data_description}. The model is trained for one epoch with a global batch size of 16 and a learning rate of 2e-6.

Stage 1 training lasted approximately 4 hours, and Stage 2 training lasted 32 hours on 3 nodes of 8 NVIDIA H100 GPUs, utilizing DeepSpeed ZeRO \citep{deepspeed2021} for distributed training. More details about our training hyperparameters are presented in \cref{table:hyperparams_general}.

Using this experimental setup, we first validate our design choices for the multi-resolution encoding by conducting multiple ablations and comparing them against baselines and alternative token reduction strategies. Training all baseline models on the full 2M instruction-tuning dataset is time-intensive, so we randomly sampled 700K samples from our supervised fine-tuning mixture to perform these ablations. All hyperparameters are kept the same as in the main experiments.

\subsection{Multiple Image Resolutions is Important}
\label{subsection:ablation_resolutions}

\begin{table}
  \caption{Ablation study results evaluating the impact of different image resolutions on model performance across multiple benchmarks. The table compares the performance of {\model} using low (L), medium (M), and high (H) resolutions individually, as well as in various combinations.}
  \label{table:ablation_image_res}
  \centering
  \small
  \begin{tabular}{lrrrrrr}
    \toprule
    \textbf{Metric}     & \textbf{L}     & \textbf{M}  & \textbf{H}  & \textbf{L + M}   & \textbf{L + H}  & \textbf{L + M + H} \\
    \midrule
    AI2D     & 60.6  & 61.8  & 60.4  & \textbf{64.5}  & 63.6  & 64.2 \\
    ScienceQA   & 76.0  & 76.2  & 76.0  & 79.2  & 79.0  & \textbf{79.7} \\
    ChartQA  & 21.6  & 48.4  & 54.1  & 52.9  & \textbf{56.2}  & \textbf{56.2} \\
    Pope-f1  & 82.2  & 87.1  & 86.0  & 87.5  & \textbf{87.7}  & \textbf{87.7} \\
    GQA      & 49.5  & 53.1  & 52.9  & 54.6  & 55.2  & \textbf{55.7} \\
    TextVQA  & 40.0  & 55.0  & 56.4  & 60.9  & 65.2  & \textbf{66.5} \\
    VizWiz   & 57.4  & 59.9  & 56.0  & 58.7  & 59.7  & \textbf{61.7} \\
    MME      & 1205.3  & 1311.6  & 1364.0  & 1227.4   & 1397.8   & \textbf{1438.9}  \\
    \bottomrule
  \end{tabular}
\end{table}

In evaluating whether high-resolution features alone are sufficient or if combining multiple resolutions leads to better performance, we trained six separate models using different combinations of image resolutions. For low resolution, we used all 576 tokens; for medium resolution, $4 \times 36$ tokens; and for high resolution, $36 \times 36$ tokens. The results, as presented in \cref{table:ablation_image_res}, provide several key insights into the role of image resolution. First, models utilizing medium or high-resolution images generally outperform those relying solely on low-resolution inputs across most benchmarks, underscoring the significance of higher resolutions in capturing fine-grained visual details. Second, combining resolutions—either low + medium or low + high—consistently exceeds the performance of individual resolutions, particularly on tasks like ChartQA and TextVQA. This demonstrates that blending global context from low-resolution images with detailed features from higher resolutions is especially effective for tasks requiring fine-grained image details. Finally, the best overall performance is achieved by integrating all three resolutions (low + medium + high), confirming that leveraging a full spectrum of image resolutions yields the highest scores across most benchmarks. These findings are consistent with conclusions drawn by models such as Llava-1.5-HD, which similarly highlight the advantages of combining multiple image resolutions \cite{llava}.

\subsection{Mean-Pooling is an Effective Token Reduction Strategy}
\label{subsection:ablation_token_reduction}

\begin{table}
    \caption{Performance comparison of multiple token reduction strategies for encoding high-resolution images against {\model}. The first model uses CLIP ViT-L/14@336px for low resolution and CLIP ViT-B/32 for medium and high resolutions, the second model is similar to {\model} but uses the IDEFICS2 perceiver resampler to reduce the number of tokens, and the third is our implementation of LLaVA-1.5-HD.}
    \label{table:architecture_ablations}
    \centering
    \begin{small}
    \begin{tabular}{@{}l r r r r@{}}
        \toprule
        \textbf{Benchmark}    & \textbf{Dual Encoder}    & \textbf{Perceiver Resampler}  & \textbf{Llava-1.5-HD}     & \textbf{Mean-Pooling ({\model})} \\
        \midrule
        AI2D   & 61.7  & 60.4  & 63.8  & \textbf{64.2}   \\
        ScienceQA   & 79.5  & 70.0  & 79.3  & \textbf{79.7}   \\
        ChartQA   & 36.6  & 48.0  & 54.0  & \textbf{56.4}   \\
        POPE-f1 & 86.2  & 84.4  & 85.7  & \textbf{87.7}  \\
        GQA & 51.8  & 53.4  & 54.1  & \textbf{55.7}  \\
        TextVQA & 48.5  & 52.6  & 64.0  & \textbf{66.5}  \\
        VizWiz & 60.4  & 56.8  & 56.1  & \textbf{61.7}  \\
        MME & 1314.9  & 1385.3  & 1414.0   & \textbf{1438.9}   \\
        \bottomrule
    \end{tabular}
    \end{small}
\end{table}

We explored several alternative token reduction strategies to compare against our mean pooling approach. The first alternative, Dual Encoder, processes low-resolution images with CLIP ViT-L/14@336px, while handling medium- and high-resolution sub-images with CLIP ViT-B/32@224px, generating 49 tokens per sub-image. Each encoder uses its own single-layer projection module, producing a total of 2,536 image tokens. As shown in \cref{table:architecture_ablations}, this approach performs worse than our mean-pooling method across most benchmarks, with significant gaps in tasks like ChartQA (36.6 vs. 56.4) and TextVQA (48.5 vs. 66.5). This demonstrates that managing the number of visual tokens by using a smaller, lower-resolution model is not optimal. Instead, leveraging a stronger encoder and compressing its output is more effective for extracting high-resolution details.

The second alternative uses a learned compression method, replacing the mean pooling layer with the IDEFICS2 perceiver resampler \citep{idefics2, perceiver}. This resampler uses 3 layers and 36 latent vectors, resulting in 2,016 tokens—matching the token count from our mean pooling approach. While a learned approach like the perceiver resampler could in principle perform better, it does not in this case, as seen in \cref{table:architecture_ablations}. For example, it underperforms on benchmarks like TextVQA (52.6 vs. 66.5) and ChartQA (48.0 vs. 56.4). This discrepancy may be due to the data scale, where the simpler mean-pooling approach proves to be more effective.

Additionally, we implemented a version of LLaVA-1.5-HD \citep{llava-next}, which processes low- and medium-resolution images using the same ViT and LLM backbone as our model but does not compress visual tokens or use high-resolution images. Our LLaVA-1.5-HD implementation generates a total of 2,880 visual tokens. Incorporating high-resolution features in LLaVA-1.5-HD improves performance over the other two baselines. However, the simple mean-pooling strategy with high-resolution image features still outperforms LLaVA-1.5-HD across all benchmarks, further reinforcing the value of generating features from zoomed-in sub-crops.

In short, the simple mean-pooling approach, when combined with high-resolution features and a powerful image encoder, consistently outperforms other token reduction strategies across benchmarks, particularly for tasks requiring fine-grained visual details.

\subsection{Disentangling Resolution and Multi-Crop Benefits}
\label{subsection:ablation_zoomin}

\begin{table}[htbp]
    \caption{Ablation study results evaluating the impact of zooming in. The table compares performance using low resolution and medium resolution, pooled down to 576 tokens, with versions starting from the low-resolution image and starting from the native-resolution image.}
  \label{table:ablation_populated_res}
  \centering
  \small
  \begin{tabular}{l p{2cm} p{3.5cm} p{3.5cm}}
    \toprule
    \textbf{Metric}     & \textbf{Low-Resolution}     & \textbf{Medium-Resolution from Low-Resolution}  & \textbf{Medium-Resolution from Native-Resolution} \\
    \midrule
    AI2D     & 60.6  & 62.9  & 61.7 \\
    ScienceQA   & 76.0  & 77.6  & 76.9 \\
    ChartQA  & 21.6  & 52.4  & 56.6 \\
    POPE     & 83.4  & 85.1  & 86.8 \\
    GQA      & 49.5  & 54.7  & 54.9 \\
    TextVQA  & 40.0  & 57.4  & 61.2 \\
    VizWiz   & 57.4  & 58.0  & 56.7 \\
    MME Perception & 1205.3 & 1398.9 & 1444.7 \\
    \bottomrule
  \end{tabular}
\end{table}

Our previous results demonstrate improved performance from our multi-resolution encoding strategy. However, it remains unclear whether these gains are primarily due to the higher image resolution preserving more information or the multi-crop approach generating separate features for each sub-image. While our method provides both benefits over a single-crop, fixed-resolution approach, we now conduct an experiment to disentangle their relative importance. Specifically, we test: 1)~the effect of generating multi-crop features from an image \textit{already downsized to low resolution}, which limits the ability to preserve extra raw image information compared to the standard single-resolution approach, and 2)~the effect of generating multi-crop features from an image that \textit{retains its native resolution}, allowing it to preserve more raw image information than both the standard low resolution approach and 1).

For the first experiment, we rescaled all images to a low resolution of $336 \times 336$, with the low-resolution performance consistent with \cref{table:ablation_image_res}. From this baseline, we conducted an experiment where we zoomed in $2 \times$, generating images of size $672 \times 672$ and producing four crops from the rescaled image. Each crop was passed through the ViT, generating 576 tokens ($24 \times 24$), which we then pooled down to 144 tokens per crop, for a total of 576 tokens across all crops. This matches the total token count of the low-resolution model. In \cref{table:ablation_populated_res}, this represents the column "Medium-Resolution from Low-Resolution", and it outperforms the "Low-Resolution" model in all benchmarks, particularly excelling in tasks like ChartQA and TextVQA, where localized information is critical. This suggests that the multi-crop approach itself, even without preserving additional raw image information, significantly contributes to improved performance, likely by enabling more focused processing of image sub-regions.

For the second experiment, without first rescaling to low resolution, we worked directly from the native-resolution image and resized it to $672 \times 672$, producing four crops from the resized image. Each crop was passed through the ViT, generating 576 tokens ($24 \times 24$), which we then pooled down to 144 tokens per crop, for a total of 576 tokens across all crops. In \cref{table:ablation_populated_res}, this represents the column "Medium-Resolution from Native-Resolution." There are two key observations here. First, as expected from previous results, this model outperforms the "Low-Resolution" baseline across all tasks. Second, it also outperforms the "Medium-Resolution from Low-Resolution" model on a majority of the tasks (5/8), highlighting the importance of preserving raw image information. However, these results indicate that most of the performance gains come from featurizing sub-crops, which remains the most important part of our approach.

\subsection{Main Results}
\label{subsection:main_results}

\begin{table*}[ht]
    \caption{Comparison of {\model} with existing VLMs of similar number of parameters across various benchmarks. The best performance is indicated in \textbf{bold} and the second-best is \underline{underlined}.}
    \label{table:main_results}
  \centering
  \tiny
  \begin{tabular}{@{}l l l c c c c c c c c c@{}}
    \toprule
    \textbf{Model}    & \textbf{Backbone}   & \textbf{\#Data} & \textbf{VQA$^{v2}$}  & \textbf{VQA$^{T}$}  & \textbf{POPE}  & \textbf{SQA} & \textbf{VizWiz} & \textbf{AI2D} & \textbf{ChartQA} & \textbf{MME} & \textbf{MMB/MMB$^{CN}$} \\
    \midrule
    InstructBLIP  & Vicuna-7B  & 130M   & -  & 50.1  & - & 60.5  & 34.5  & - & -  & -  & 36.0/23.7 \\
    Qwen-VL-Chat   & Qwen-7B    & 1.4B   & 78.2  & 61.5  & - & 68.2  & 38.9  & 62.3 & \underline{65.7} & 1487.5 & 60.6/56.7 \\
    LLaVA-1.5     & Vicuna-7B  & 1.2M   & 78.5  & 58.2  & 85.9 & 66.8 & 50.0  & 54.8 & 18.2  & 1510.7  & 63.4/58.3   \\
    VILA           & Llama2-7B  & 61M   & 79.9  & 64.4  & 85.5 & 68.2  & 57.8  & - & -  & 1533.0  & 68.9/61.7 \\
    LLaVA-NeXT     & Vicuna-7B  & 1.2M   & \underline{81.8}  & 64.9  & 86.5 & 70.1  & 57.6  & \underline{66.6} & 54.8  & 1519.0  & 67.4/60.6 \\
    MM1-7B-Chat    & MM1-7B     & >2B   & \textbf{82.3}  & \underline{72.8}  & 86.6 & 72.6  & 45.3  & - & -  & 1529.3  & \underline{72.3}/- \\
    mPLUG-Owl2     & Llama2-7B  & 401M   & 79.4  & 58.2  & 86.2 & 68.7  & 54.5  & - & -  & 1450.2  & 63.5/- \\
    Monkey         & Qwen-7B    & 1B    & 80.3  & - & 67.6  & 69.4 & \textbf{61.2}   & 62.6 & 65.1  & -  & - \\
    SPHINX         & Llama2-7B  & 1B    & 78.1  & 51.6  & 80.7  & 69.3  & 39.9 & - & -  & 1476.1  & 66.9/56.2 \\
    SPHINX-2k      & Llama2-7B  & 1B    & 80.7  & 61.2  & 87.2 & 70.6  & 44.9  & - & -  & 1470.7  & 65.9/57.9 \\
    ShareGPT4V-7B  & Vicuna-7B  & 1.8M   & 80.6  & -  & - & 68.4  & 57.2  & - & -  & \textbf{1567.4}  & 68.8/62.2 \\
    VisionLLM v2-chat & Vicuna-7B  & 22M   & 81.4  & 66.3  & \underline{87.5} & \textbf{94.4}  & 54.6  & - & -  & 1512.5  & \textbf{77.1}/\textbf{67.6} \\
    InternVL-7B    & Vicuna-7B  & >28.7B    & 79.3  & 57.0  & 86.4 & 66.2  & 52.5  & - & -  & 1525.1  & 64.6/57.6 \\
    \midrule
    {\model} (Ours) & Llama3-8B  & 2.9M  & 81.0  & \textbf{73.6}  & \textbf{87.9} & \underline{79.5}  & \underline{59.0}  & \textbf{67.9}  & \textbf{71.2}  & \underline{1538.1}  & 71.9/\underline{66.1} \\
    \bottomrule
  \end{tabular}
\end{table*}

Following our validation of design choices in the previous experiments, we now train our model on a scaled-up instruction-tuning dataset of 2 million samples and compare {\model} against other open-source VLMs. The results, presented in \cref{table:main_results}, evaluate the models across ten established benchmarks, including general-domain visual question answering datasets (ScienceQA, VQA$^{v2}$, VizWiz; \citealt{scienceqa, antol2015vqa, gurari2018vizwiz}), chart interpretation and OCR-based VQA datasets (ChartQA and TextVQA; \citealt{chartqa, singh2019towards}), hallucination assessment datasets (POPE; \citealt{pope}), and other standard benchmarks such as AI2D \citep{ai2d}, MME \citep{fu2023mme}, MMB \citep{liu2023mmbench}, and MMB$^{CN}$ (the Chinese-language version of MMB).

One of the key areas where {\model} excels is in tasks requiring fine-grained visual understanding, such as TextVQA and ChartQA. For instance, {\model} achieves a score of 73.6 on TextVQA and 71.2 on ChartQA, outperforming all other models in the table. By comparison, Qwen-VL-Chat \citep{qwenvl}, trained on over 400 times more data, scores only 61.5 on TextVQA and 65.7 on ChartQA. This result aligns with previous research \citep{beyer2024paligemma}, which emphasizes the importance of high-resolution images for tasks involving intricate visual details, such as text recognition and chart interpretation.

Additionally, {\model} achieves the best performance on POPE-f1 (87.9) and ranks second-best on VizWiz (59.0), MME (1538.1), ScienceQA (79.5), and MMB$^{CN}$ (66.1). The models that outperform {\model} on certain benchmarks, such as MM1-7B-Chat \citep{mm1} and Monkey \citep{li2024monkey}, are trained on significantly larger datasets with over 1 billion samples.

As shown in \cref{table:main_results_13B}, {\model} also competes strongly against 13B-17B models across various benchmarks. It outperforms all 13B models on TextVQA, ChartQA, and MMB$^{CN}$, while achieving second-best performance on POPE, ScienceQA, VizWiz, AI2D, and MMB, competing against powerful models such as CogVLM-17B-Chat \citep{wang2023cogvlm}. This underscores {\model}'s efficiency in leveraging high-resolution, zoomed-in image features and a powerful visual encoder.

It is important to note that these comparisons involve several differences, including variations in training stages, data, and underlying ViTs and LMs. Despite these differences, {\model} remains competitive, particularly on text and chart interpretation tasks, even though it uses significantly less data than many other models. This highlights the utility of our zoomed-in encoding strategy and suggests that its effectiveness would likely increase with additional training data.

\subsection{Biomedical Domain Adaptation}
\label{subsection_biomedical_domain_adapt}

\begin{figure}[t]
    \centering
    \includegraphics[width=\textwidth]{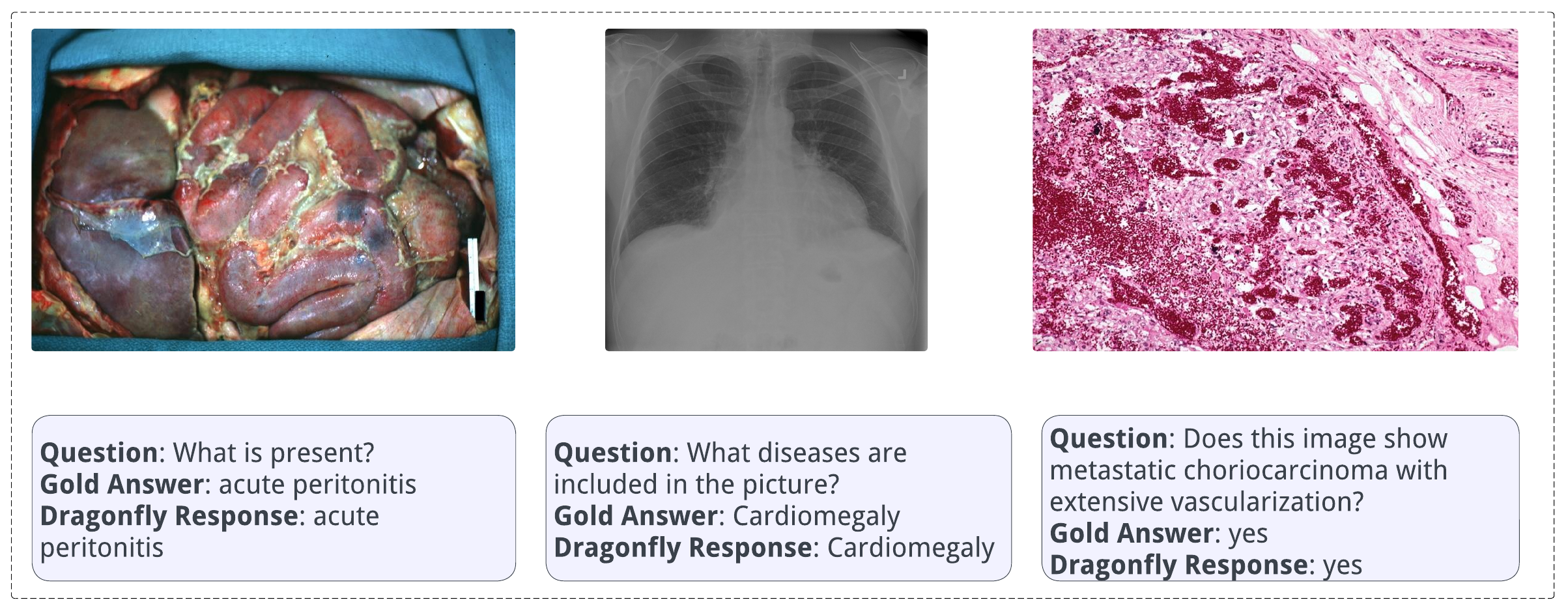}
    \caption{Examples of biomedical VQA. The figure shows three questions along with their gold standard answers and the corresponding responses from the {\modelbiomed} model.}
    \label{fig:biomed_vqa_examples}
\end{figure}

This section outlines our approach to adapting the model for the biomedical domain, enabling its evaluation across a range of specialized and challenging tasks.  Starting with a model checkpoint instruction-tuned on our general-domain dataset, we implemented a three-step training process tailored specifically for the biomedical domain to create {\modelbiomed}.

The first stage involved tuning the projection layer and vision encoder, which is critical given the limited exposure of the standard CLIP encoder to biomedical images. The training dataset for this phase primarily comprised short caption datasets from sources like LLaVA-Med \citep{li2024llava}, OpenPath \citep{huang2023leveraging}, and MedICaT \citep{subramanian2020medicat}, supplemented by general domain datasets from LLaVA-Pretrain \citep{llava}. This phase included approximately 1.16M image-text pairs, split roughly evenly between the general and biomedical domains. Stage 1 took approximately 24 hours to train on 8 NVIDIA H100 GPUs.

In the second stage, we jointly trained the vision encoder, language model, and projection layer. We used a diverse set of datasets, including LLaVA-Med-Instruct \citep{li2024llava}, MIMIC-III-CXR \citep{johnson2019mimic}, OpenPath \citep{huang2023leveraging}, ROCO \citep{pelka2018radiology}, Kaggle DR, and DDR \citep{li2019diagnostic}. Additionally, we included training sets from benchmark datasets such as VQA-RAD \citep{lau2018dataset}, SLAKE \citep{liu2021slake}, Path-VQA \citep{he2020pathvqa}, IU X-Ray, and Peir Gross \citep{demner2016preparing}. The dataset totaled 723K image-text pairs, with approximately 15\% from the general domain and 85\% from the biomedical domain. General domain datasets included SVIT \citep{zhao2023svit}, ShareGPT4V \citep{sharegpt4v}, and ArXivCap \citep{li2024multimodal}. Stage 2 took about 30 hours on 8 NVIDIA H100 GPUs.

The final stage involved fine-tuning using combined training datasets from our benchmark tasks: VQA-RAD, SLAKE, Path-VQA, IU X-Ray, Peir Gross, and subsets of ROCO and MIMIC-CXR, totaling 50K image-text pairs. We fine-tuned a single model end-to-end on this aggregated training data to optimize performance across all tasks simultaneously. Stage 3 required approximately 4 hours of training on 8 NVIDIA H100 GPUs.

\begin{table}[htbp]
    \centering
    \small
    \caption{Medical image captioning and clinical report generation evaluation results. For MIMIC-CXR, we specifically focus on generating the findings section of the radiology report.}
    \label{tab:caption_biomed_results}
    \begin{tabular}{@{}ccccc@{}}
        \toprule
        \textbf{Dataset} & \textbf{Metric} & \textbf{BiomedGPT} & \textbf{SOTA} & \textbf{{\modelbiomed} (Ours)} \\
        \midrule
        \multirow{3}{*}{IU X-Ray}   & ROUGE-L & 28.5 & 44.8 \citep{zhou2021visual} & 29.1 \\
                                    & METEOR  & 12.9 & 24.2 \citep{huang2023kiut}  & \textbf{30.5} \\
                                    & CIDEr   & 40.1 & 43.5 \citep{wang2023met}    & \textbf{61.7} \\
        \midrule
        \multirow{3}{*}{Peir Gross} & ROUGE-L & 36.0 & 36.0 \citep{zhang2023biomedgpt} & \textbf{42.0} \\
                                    & METEOR  & 15.4 & 15.4 \citep{zhang2023biomedgpt} & \textbf{40.2} \\
                                    & CIDEr   & 122.7 & 122.7 \citep{zhang2023biomedgpt} & \textbf{198.5} \\
        \midrule
        \multirow{3}{*}{ROCO}       & ROUGE-L & 18.2 & 18.2 \citep{zhang2023biomedgpt} & \textbf{19.2} \\
                                    & METEOR  & 7.8  & 7.8  \citep{zhang2023biomedgpt} & \textbf{15.5} \\
                                    & CIDEr   & 24.2 & 24.2 \citep{zhang2023biomedgpt} & \textbf{45.2} \\
        \midrule
        \multirow{3}{*}{MIMIC-CXR}  & ROUGE-L & 23.8  & 33.5 \citep{zhou2021visual} & 25.2 \\
                                    & METEOR  & 14.2  & 19.0 \citep{zhou2021visual} & \textbf{23.6} \\
                                    & CIDEr   & 14.7  & \textbf{50.9} \citep{miura2020improving} & \textbf{50.9} \\
        \bottomrule
    \end{tabular}
\end{table}

\begin{table}[htbp]
\centering
\scriptsize
\caption{Biomedical VQA evaluation results.}
\label{tab:vqa_biomed_results}
\begin{tabular}{@{}cccccc@{}}
\toprule
\textbf{Dataset} & \textbf{Metric} & \textbf{LLaVA-Med} & \textbf{Med-Gemini} & \textbf{SOTA} & \textbf{\modelbiomed (Ours)} \\
\midrule
\multirow{2}{*}{VQA-RAD} & Acc (closed) & 84.2 & 69.7 & 87.1 \citep{tanwani2022repsnet} & 78.1 \\
                         & Token F1     & - & 50.1 & 62.1 \citep{tu2024towards} & 61.4 \\
\midrule
\multirow{2}{*}{SLAKE}   & Acc (closed) & 83.2 & 84.8 & \textbf{91.6} \citep{yuan2023ramm} & \textbf{91.6} \\
                         & Token F1     & - & 75.8 & \textbf{89.3} \citep{tu2024towards} & \textbf{89.3} \\
\midrule
\multirow{2}{*}{Path-VQA}& Acc (closed) & 91.7 & 83.3 & 91.7 \citep{li2024llava} & 90.6 \\
                         & Token F1     & - & 58.7 & 62.7 \citep{tu2024towards} & \textbf{67.1} \\
\bottomrule
\end{tabular}
\end{table}

The results, as reported in \cref{tab:caption_biomed_results} and \cref{tab:vqa_biomed_results}, are based on this fine-tuned model and evaluated against the official held-out test sets of the respective benchmarks (details of the biomedical benchmarks are provided in \cref{app:biomedical-benchmark}). For VQA tasks, we use accuracy and token-level F1 \citep{tu2024towards}, while for image captioning and radiology report generation tasks, we use standard metrics such as ROUGE-L \citep{lin2004rouge}, METEOR \citep{banerjee2005meteor}, and CIDEr \citep{vedantam2015cider}. These metrics evaluate the fluency of text and the recognition of synonyms and word stems, with CIDEr specifically tailored for assessing text descriptions of images.

{\modelbiomed} achieves strong performance across multiple benchmarks. On the image captioning task, {\modelbiomed} delivers state-of-the-art or competitive results on several metrics across these datasets. Notably, on the Peir Gross and ROCO datasets, {\modelbiomed} outperforms existing methods on all three metrics: ROUGE-L, METEOR, and CIDEr. On the other two captioning benchmarks (IU X-Ray and MIMIC-CXR), {\modelbiomed} achieves state-of-the-art performance on two out of three evaluation metrics. 

For VQA tasks, {\modelbiomed} attains an accuracy of 91.6\% and a token F1 score of 89.3\% on the SLAKE dataset, matching the current state-of-the-art. Similarly, on Path-VQA, {\modelbiomed} sets a new state-of-the-art performance with a token F1 score of 67.1, surpassing the much larger Med-PaLM-M model, which scores 62.7. Additionally, {\modelbiomed} consistently outperforms Med-Gemini, a significantly larger model, on all VQA tasks. These results further highlight the fine-grained understanding and reasoning capabilities of the {\modelbiomed} architecture. \cref{fig:biomed_vqa_examples} presents a few examples from our evaluation tasks, along with {\modelbiomed}'s responses.

%

\section{Discussion and Conclusion}
\label{section_conclusion}

High-resolution image inputs help capture fine-grained visual details, which are critical for tasks such as OCR and reading charts. Our study demonstrates that leveraging powerful vision encoders and pushing image resolutions beyond native sizes enhances the model's ability to identify subtle visual cues. Zooming in beyond native resolution allows the model to capture fine-grained details that might otherwise be missed, particularly in small objects, dense text, and chart details. We show that a multi-resolution encoding strategy, paired with a simple mean pooling compression approach, provides an effective and computationally efficient solution, preserving both global context and fine details. {\model} even surpasses larger models in several benchmarks while utilizing fewer tokens and less data. 

Despite the strong performance of {\model}, there are several limitations to our approach. First, while we've demonstrated competitive performance using much smaller datasets than other models, further investigation is needed to explore the potential for even greater improvements as the approach scales with larger supervised fine-tuning datasets. Second, while the increased resolution and multiple image crops enhance the model's visual understanding, they come at the cost of higher computational demands in the vision encoder. Having said that, by applying mean pooling to compress sub-image representations, we ensure that the context length passed to the LLM remains manageable and mitigates the impact of these additional FLOPs. 

Interestingly, the strong performance of our simple approach—zooming in beyond native resolution and mean pooling the tokens—highlights a broader issue: the fixed-resolution approach of current vision transformers is inherently limiting. While multi-crop strategies offer some improvement, they introduce complexity and increased computational demands. Moving forward, VLMs should adopt native-resolution architectures that can process images at various scales in a single pass, preserving all the information without requiring multiple crops. Additionally, improved training strategies are needed to ensure that models retain the same level of detail as if magnified sub-crops were processed individually.





\bibliography{main}
\bibliographystyle{styles/iclr2025_conference}

\appendix
\section{General Domain Training Data Description}
\label{app:section_general_data_description}

We curated a vision instruction-tuning dataset using samples from ShareGPT4V \citep{sharegpt4v}, ALLaVA \citep{allava}, SVIT \citep{svit}, and selected tasks from Cauldron \citep{idefics2}. Initially, we combined the samples from these four sources, resulting in nearly 9 million data points. Through experimentation with the training data, we derived several key insights:

\begin{itemize}[leftmargin=1.5em]
    \item Increasing the number of training samples during visual instruction tuning improves the model’s performance on commonsense reasoning tasks but also increases the likelihood of hallucination. To mitigate this, the model benefits from training on specialized data.
    \item Deduplicating the training samples is crucial. Duplicate samples can introduce bias during training, negatively impacting model performance.
    \item Question-answering data enhances benchmark performance but can reduce the detail and length of generated text.
\end{itemize}

Based on these insights, we first deduplicated the image-instruction pairs. Since SVIT and ShareGPT4V share the same image set, and SVIT generates multiple instructions per image, we randomly selected eight instructions per image to scale the dataset. The Cauldron dataset, a vast collection of 50 high-quality datasets converted to user/assistant format, included some datasets related to math or coding, which caused misalignment during training. As a result, we excluded five datasets from Cauldron. After processing and deduplication, our final training set contained 2 million image-instruction pairs. Additionally, we included text-only data from OpenHermes and MathInstruct to maintain the model’s zero-shot capabilities.

\begin{figure}[htbp]
    \centering
    \includegraphics[width=0.8\textwidth]{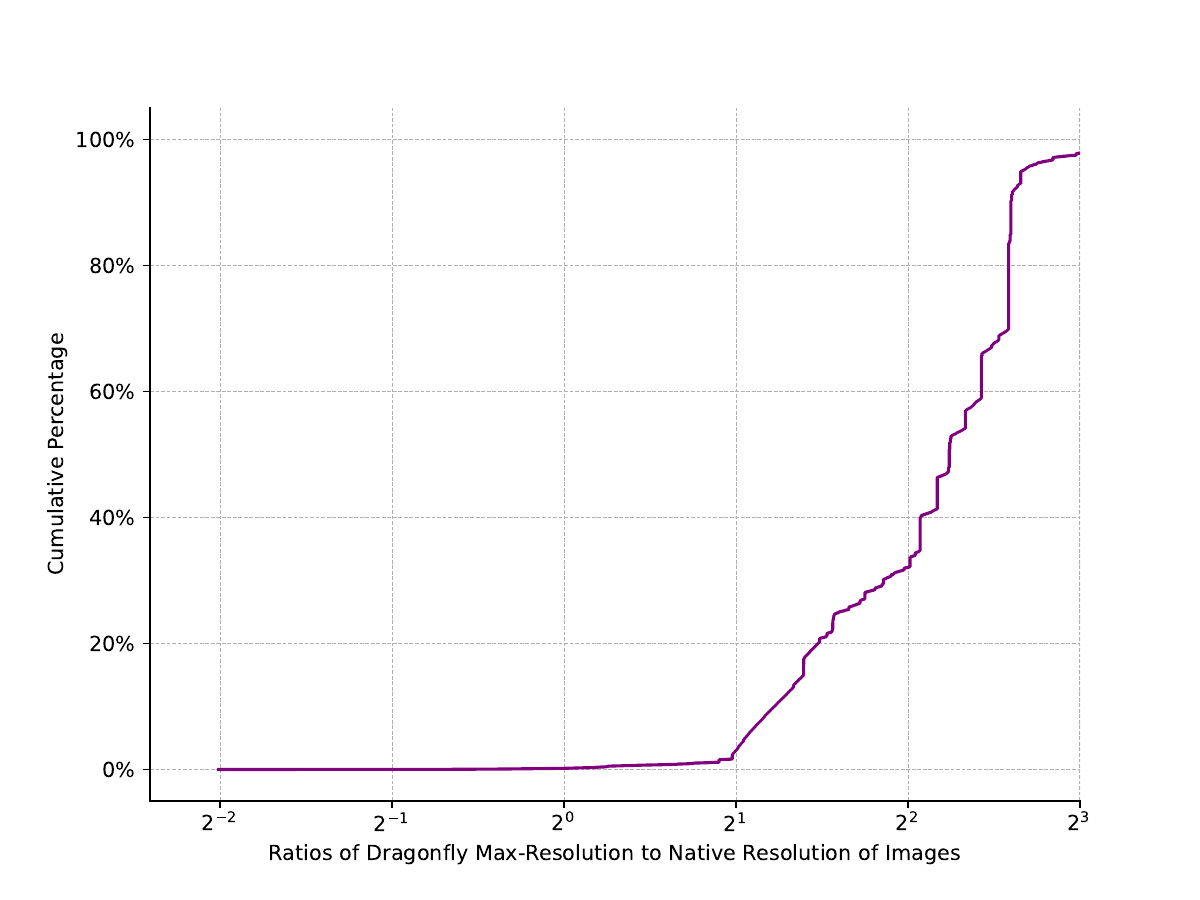}
    \caption{Ratio of maximum resolution of our high resolution image to the native resolution of the original image. We used all of our training dataset to calculate this ratio, which comprised data from multiple different sources and tasks. First, we matched each image into one of the aspect ratios with the algorithm mentioned in \ref{subsection:experimental_setup}. Then, we calculated the ratio between the longest dimension in our max-resolution to the longest dimension in the native resolution of the image. From the plot, we can see that 65\% of the images in our training cohort are zoomed-in by at least 4x the native resolution.}
    \label{fig:max_res_kde}
\end{figure}

\begin{table}[ht]
\centering
\scriptsize
\caption{Summary of the evaluation benchmarks for general domain.}
\begin{tabular}{@{}lllll@{}}
\toprule
\textbf{Task}              & \textbf{Dataset} & \textbf{Description}                                          & \textbf{Split}         & \textbf{Metrics} \\ 
\midrule
\textbf{General VQA}       & VQA$^{v2}$       & VQA on natural images.                                        & test-dev               & Accuracy (↑)    \\
                           & ScienceQA        & Multi-choice VQA on a diverse set of science topics.           & test                   & Accuracy (↑)     \\
                           & VizWiz           & VQA on images taken by visually impaired users.                & test                   & Accuracy (↑)     \\
                           & AI2D             & VQA on diagrams and other artificial images.                   & test                   & Accuracy (↑)     \\
\midrule
\textbf{Text-oriented VQA} & TextVQA          & VQA on natural images containing text.                         & val                    & Exact Match (↑)    \\
                           & ChartQA          & VQA on various types of charts and graphs.                     & test                   & Accuracy (↑)     \\
\midrule
\textbf{LVLM Benchmarks}   & MMBench          & Multi-choice VQA on a diverse set of topics.                   & test                   & Accuracy (↑)     \\
                           & MMBench$^{CN}$   & Multi-choice VQA on a diverse set of topics in Chinese.         & test                   & Accuracy (↑)     \\
                           & POPE             & Multi-choice VQA for testing hallucinations.                   & overall                & Accuracy (↑)     \\
                           & MME              & Multi-modal evaluation benchmark for general VQA abilities.     & test                   & Accuracy (↑)     \\
\bottomrule
\end{tabular}
\end{table}

\begin{table}[htbp]
    \centering
    \caption{Selected hyperparameters for Stage 1 and Stage 2 training of {\model}.}
    \label{table:hyperparams_general}
    \begin{tabular}{@{}lcc@{}}
        \toprule
        Hyperparameter                & Stage 1         & Stage 2         \\
        \midrule
        Batch Size                    & 64               & 16               \\
        Learning Rate                 & 2e-5            & 2e-6            \\
        LR Scheduler                  & cosine          & cosine          \\
        Warmup Steps Ratio            & 0.01            & 0.01            \\
        Max Sequence Length           & 4096            & 4096            \\
        Tune Projection Layer         & \checkmark      & \checkmark      \\
        Tune Vision Encoder           & \texttimes      & \checkmark      \\
        Tune LLM                      & \texttimes      & \checkmark      \\
        \bottomrule
    \end{tabular}
\end{table}

\begin{table*}[ht]
  \caption{Model architectures and data usage details for our model and baseline models.}
  \label{table:model_architecture}
  \centering
  \small
  \begin{tabular}{@{}l l l c c c@{}}
    \toprule
    \textbf{Model} & \textbf{LLM Backbone} & \textbf{Vision Base} & \textbf{\#Data} & \textbf{Max res} \\
    \midrule
    InstructBLIP \citep{dai2023instructblip}     & Vicuna-7B    & CLIP-g/14      & 130M     & 224$\times$224     \\
    Qwen-VL-Chat \citep{qwenvl}    & Qwen-7B      & CLIP-bigG   & 1.4B     & 448$\times$448     \\
    LLaVA-1.5 \citep{liu2024improved}      & Vicuna-7B    & CLIP-L/14 & 1.2M     & 336$\times$336     \\
    VILA \citep{lin2024vila}            & Llama2-7B    & CLIP-L/14      & 51M     & 364$\times$364 \\
    LLaVA-NeXT \citep{liu2024llava}     & Vicuna-7B    & CLIP-L/14 & 1.2M     & 672$\times$672     \\
    MM1-7B-Chat \citep{mm1}    & MM1-7B       & CLIP-H        & >2B      & 378$\times$378     \\
    mPLUG-Owl2 \citep{ye2024mplug}     & Llama2-7B    &  CLIP-L/14   & 401M    & 448$\times$448     \\
    Monkey \citep{li2024monkey}          & Qwen-7B      & CLIP-BigG        & 1B      & 896$\times$1344     \\
    SPHINX \citep{lin2023sphinx}          & Llama2-7B    & Mixed Encoders        & 1B     & 448$\times$448     \\
    SPHINX-2k   \citep{lin2023sphinx}     & Llama2-7B    & Mixed Encoders      & 1B     & 762$\times$762     \\
    ShareGPT4V-7B \citep{sharegpt4v}   & Vicuna-7B    & CLIP-L/14       & 1.8M    & 336$\times$336     \\
    VisionLLM v2-chat \citep{wu2024visionllm} & Vicuna-7B  & CLIP-L/14 & 22M    & 336$\times$336      \\
    InternVL-7B \citep{chen2024internvl}    & Vicuna-7B    & InternViT-6B   & >28.7B   & 224$\times$224 \\
    \midrule
    InstructBLIP \citep{dai2023instructblip}   & Vicuna-13B   & CLIP-g/14      & 130M    & 224$\times$224      \\
    LLaVA-1.5 \citep{liu2024improved}      & Vicuna-13B   & CLIP-L/14 & 1.2M    & 336$\times$336     \\
    VILA \citep{lin2024vila}  & Llama2-13B   & CLIP-L/14      & 51M    & 364$\times$364     \\
    LLaVA-NeXT \citep{liu2024llava}      & Vicuna-13B   & CLIP-L/14 & 1.2M    & 672$\times$672     \\
    LLaVA-UHD \citep{llava-uhd}      & Vicuna-13B   & CLIP-L/14 & 1.2M    & 672$\times$1008     \\
    InternVL-13B \citep{chen2024internvl}  & Vicuna-13B   & InternViT-6B   & >28.7B    & 364$\times$364 \\
    CogVLM-17B-Chat  \citep{wang2023cogvlm}   & Vicuna-7B    & EVA2-CLIP-E & >1.5B    & 490$\times$490    \\
\midrule
\model~ (Ours)  & Llama3-8B    & ViT-L/14        & 2.9M   & 2016$\times$2016 or \\
                &&&&1008$\times$4032\\ 
\bottomrule
  \end{tabular}
\end{table*}

\begin{table*}[ht]
    \caption{Comparison between {\model} and existing LMMs across various benchmarks. \textbf{Bold} numbers indicate the best performance among all the 13B models, while \underline{underlined} numbers represent the second-best performance.}
  \label{table:main_results_13B}
  \centering
  \tiny
  \begin{tabular}{@{}l l l c c c c c c c c c@{}}
    \toprule
    \textbf{Model}    & \textbf{Backbone}   & \textbf{\#Data} & \textbf{VQA$^{v2}$}  & \textbf{VQA$^{T}$}  & \textbf{POPE}  & \textbf{SQA} & \textbf{VizWiz} & \textbf{AI2D} & \textbf{ChartQA} & \textbf{MME} & \textbf{MMB/MMB$^{CN}$} \\
    \midrule
    InstructBLIP   & Vicuna-13B & 130M  & -  & 50.7  & 78.9 & 63.1  & 33.4  & - & -  & 1212.8  & - \\
    LLaVA-1.5      & Vicuna-13B & 1.2M  & 80.0  & 61.3  & 85.9 & 71.6  & 53.6  & 59.5 & 18.2  & 1531.3  & 66.9/63.6   \\
    VILA           & Llama2-13B & 51M   & 80.8  & 66.6  & 84.2 & 73.7  & \textbf{60.6}  & - & -  & \underline{1570.1}  & 70.3/64.3 \\
    LLaVA-NeXT     & Vicuna-13B & 1.2M  & \textbf{82.8}  & 67.1  & 86.2 & 73.6  & \textbf{60.6}  & \textbf{70.0} & \underline{62.2}  & \textbf{1575.0}  & 70.0/64.4 \\
    LLaVA-UHD      & Vicuna-13B & 1.2M  & \underline{81.7}  & 67.7  & \textbf{89.1} & 72.0  & 56.1  & - & -  & 1535.0  & 68.0/\underline{64.8} \\
    InternVL-13B   & Vicuna-13B & 6B    & 80.2  & 58.7  & 87.1 & 70.1  & 54.6  & - & -  & 1546.9  & 66.5/61.9 \\
    CogVLM-13B-Chat    & Vicuna-7B  & >1.5B  & 82.3  & \underline{70.4}  & 87.9 & \textbf{91.2}  & -  & - & -  & -  & \textbf{77.6}/- \\
    \midrule
    \model~ (Ours) & Llama3-8B  & 2.9M  & 81.0  & \textbf{73.6}  & \underline{87.9} & \underline{79.5}  & \underline{59.0}  & \underline{67.9}  & \textbf{71.2}  & 1538.1  & \underline{71.9}/\textbf{66.1} \\
    \bottomrule
  \end{tabular}
\end{table*}

\section{Biomedical Training Data Description}
\label{app:section_biomed_data_description}

Many public datasets were used in the training and evaluation of \model. All datasets were de-identified. Open datasets were used following their existing licenses. 

\subsection{LLaVA-Med}

LLaVA-Med is a dataset for instruction-following tasks involving multi-round conversations about biomedical images, generated using the language-only model GPT-4 (\cite{li2024llava}). Specifically, the model is prompted to generate questions and answers in multi-round formats based on an image caption, as if it could view the image itself. To assemble the image captions and their contexts, LLaVA-Med utilizes PMC-15M (\cite{zhang2023biomedclip}) to select images that contain a single plot. From these, it samples 60,000 image-text pairs from the five most prevalent imaging modalities: CXR (chest X-ray), CT (computed tomography), MRI (magnetic resonance imaging), histopathology, and gross pathology. The dataset also extracts sentences referencing the image from the original PubMed articles to provide additional context to the captions. LLaVA-Med offers two primary versions of the dataset: (i) 60K-IM, which includes inline mentions as context, and (ii) 60K, a similar-sized dataset that excludes inline mentions in its self-instruct generations. Furthermore, a supplementary dataset of 500,000 image-caption pairs is available for alignment purposes. Data link: \url{https://github.com/microsoft/LLaVA-Med}

\subsection{Medicat}

Medicat (\cite{subramanian2020medicat}) is a dataset of medical figures, captions, subfigures/subcaptions, and inline references that enables the study of these figures in context. It consists of 217,000 images from 131,000 open-access PubMed Central and includes captions, inline references for 74\% of figures, and manually annotated subfigures and subcaptions for a subset of figures. Data link: \url{https://github.com/allenai/medicat}.

\subsection{MIMIC-III-CXR}

The MIMIC-III-CXR dataset (\cite{johnson2019mimic}) is a substantial publicly available collection of chest radiographs, containing 377,110 images derived from 227,827 imaging studies conducted at the Beth Israel Deaconess Medical Center from 2011 to 2016. Each image in the dataset is paired with structured labels extracted from free-text radiology reports. The dataset is organized into training, validation, and testing subsets, with 368,960 images allocated for training, 2,991 for validation, and 5,159 for testing. To ensure patient confidentiality, all images have been de-identified. Data link: \url{https://physionet.org/content/mimic-cxr-jpg/2.1.0/}

\subsection{Openpath}

OpenPath dataset is an expansive collection of 208,414 pathology image-text pairs, making it the largest publicly available pathology image dataset annotated with text descriptions (\cite{huang2023leveraging}). This dataset was meticulously curated using popular pathology-related hashtags recommended by the United States and Canadian Academy for Pathology (USCAP) and the Pathology Hashtag Ontology projects. It spans images gathered from Twitter and other internet sites, including the LAION dataset, collected between March 21, 2006, and November 15, 2022. The dataset consists of three main components: (1) Tweets, with 116,504 image-text pairs; (2) Replies, comprising 59,869 pairs from highly liked responses; and (3) PathLAION, which adds 32,041 pairs from broader internet sources. Data link: \url{https://github.com/PathologyFoundation/plip}.

\subsection{Kaggle DR (Diabetic Retinopathy)}

The Kaggle website organized a DR detection challenge in 2015 \cite{li2019diagnostic}. The California Healthcare Foundation sponsored the competition. The Kaggle DR dataset consists of 88,702 color fundus images, including 35,126 samples for training and 53,576 samples for testing. Different devices captured the images under various conditions (e.g., resolutions) at multiple primary care sites throughout California and elsewhere. For each subject, two images of the left and right eyes were collected with the same resolution. Clinicians rate each image for the presence of DR on a scale of 0–4 according to the ETDRS scale. Data link: \url{https://www.kaggle.com/c/diabetic-retinopathy-detection}.

\subsection{DDR}

DDR is a diabetic retinopathy dataset (\cite{li2019diagnostic}) that comprises 13,673 color fundus images collected from 147 hospitals across 23 provinces in China between 2016 and 2018, ensuring a broad demographic spread by including images from patients aged 1 to 100, averaging 54.13 years, and almost evenly split between males (48.23\%) and females (51.77\%). These images, derived from 9,598 patients and captured using 42 types of fundus cameras, adhere to stringent photographic standards to ensure clarity and appropriate exposure, focusing on crucial retinal structures and lesions. All images have been desensitized for widespread usage and graded for diabetic retinopathy (DR) severity by seven trained graders using the International Classification of Diabetic Retinopathy, supplemented by consensus and consultation with experienced specialists where necessary. Data link: \url{https://github.com/nkicsl/DDR-dataset}.

\subsection{ROCO}

The Radiology Objects in Context (ROCO) dataset is a comprehensive collection of over 81,000 radiology images derived from PubMedCentral's open-access biomedical literature (\cite{pelka2018radiology}). The dataset focuses on analyzing visual elements and semantic relationships within radiological imagery. It includes a variety of medical imaging modalities such as Computer Tomography (CT), Ultrasound, X-ray, Fluoroscopy, Positron Emission Tomography (PET), Mammography, Magnetic Resonance Imaging (MRI), and Angiography. Each image is accompanied by detailed metadata, including captions, keywords, and identifiers from the Unified Medical Language System (UMLS). The ROCO dataset also features an out-of-class set of approximately 6,000 images, ranging from synthetic radiology figures to digital art, to aid in improving prediction and classification tasks. The dataset is split into training, validation, and test sets with 70,308, 8,782, and 8,786 images, respectively.

\subsection{VQA-RAD}

The VQA-RAD dataset (\cite{lau2018dataset}) contains 314 radiology images and 2,244 question-answer pairs obtained from CT, MRI, and X-ray examinations, covering three anatomical regions: the head, abdomen, and chest. It features a diverse range of question styles, categorized into 11 types: modality, plane, organ system, abnormalities, etc. Among these, 58\% of the question-answer pairs are closed-ended (yes/no), with the remaining 42\% being open-ended. The dataset is segmented into a training set of 1,790 QA pairs and a testing set of 451 QA pairs. Our model was trained on the official training set and evaluated on the official test set. Data link: \url{https://huggingface.co/datasets/flaviagiammarino/vqa-rad}.

\subsection{SLAKE}

The Slake-VQA dataset, annotated by expert physicians (\cite{liu2021slake}), is a comprehensive bilingual (English and Chinese) VQA dataset. It includes 642 images and 14,028 question-answer pairs across three imaging modalities: CXR, CT, and MRI. This dataset spans various radiological areas, covering body regions such as the brain, neck, chest, abdomen, and pelvic cavity. It contains 9,849 VQA samples designated for training, 2,109 for validation, and 2,070 for testing. The questions vary widely, featuring both open-ended (free-form) and closed-ended (yes/no) types that assess different image characteristics like plane, quality, position, organ, abnormality, size, color, shape, and pertinent medical knowledge. We utilized only the English-language examples from the official dataset divisions, comprising 4,919 training, 1,053 validation, and 1,061 test examples. Our model was trained on the official training set and evaluated on the official test set. Data link: \url{https://www.med-vqa.com/slake/}

\subsection{Path-VQA}

This dataset comprises question-answer pairs relating to pathology images (\cite{he2020pathvqa}). It encompasses a variety of question formats, including open-ended and closed-ended (yes/no) questions. The dataset is constructed through automated techniques and draws from two open-access pathology textbooks and a digital library. It encompasses a total of 32,632 question-answer pairs derived from 4,289 images. The dataset is partitioned into official training, validation, and test subsets, containing 19,654, 6,259, and 6,719 QA pairs, respectively. Our model was trained on the official training set and evaluated on the official test set. Data link: \url{https://github.com/UCSD-AI4H/PathVQA/tree/master/data}

\subsection{IU X-ray}

The IU X-ray dataset, detailed in \cite{demner2016preparing}, is available through the Open Access Biomedical Image Search Engine (OpenI). This collection includes radiological exams or cases, each associated with one or more images, a radiology report, and two sets of tags. The reports consist of four sections: Comparison, Indication, Findings, and Impression, with the latter two sections useful for image captioning. The dataset features two types of tags: MTI tags derived automatically from the report text by the Medical Text Indexer and manual tags assigned by two trained coders. Overall, it comprises 3,955 reports and 7,470 frontal and lateral X-ray images. The dataset is divided into 6,698 samples in the training set and 745 samples in the test set. Data link: \url{https://github.com/nlpaueb/bioCaption}

\subsection{Peir Gross}

The Peir Gross dataset, initially utilized for captioning in research by \cite{jing2017automatic}, features photographs from medical cases sourced from the Pathology Education Informational Resource (PEIR) digital library intended for educational purposes in pathology. This dataset includes 7,443 images from the Gross collections across 21 pathology sub-categories in PEIR, with each image paired with a descriptive single-sentence caption. It is organized into two subsets: 5,172 images for training and 1,289 for testing. Data link: \url{https://github.com/nlpaueb/bioCaption}

\section{Biomedical Benchmarks}
\label{app:biomedical-benchmark}

The details of our evaluation benchmarks are discussed in Section \ref{app:section_biomed_data_description}. A benchmark summary table is also included in \ref{tab:multimedbench}.

\begin{table}[htbp]
\centering
\caption{Summary of the biomedical evaluation benchmark, which includes vision question answering, image captioning, and report generation across radiology and pathology modalities. We finetuned the model using a subset of the official training set and evaluated it on the official testing set. It should be noted that for MIMIC-CXR and ROCO, we utilized only a portion of the training dataset. Furthermore, for MIMIC-CXR, we selected only those subsets of the test set, including a findings section.}
\label{tab:multimedbench}
\begin{tabular}{@{}p{3cm}p{2.0cm}p{2cm}cc@{}}
\toprule
\textbf{Task Type} & \textbf{Modality} & \textbf{Dataset} & \multicolumn{2}{c}{\textbf{Split}} \\
\cmidrule(lr){4-5}
& & & \textbf{Train} & \textbf{Test} \\
\midrule
\multirow{3}{*}{\parbox{3cm}{Visual Question \\Answering}} & Radiology & VQA-RAD & 1,790 & 451 \\
 & Radiology & Slake-VQA & 4,919 & 1,053 \\
 & Pathology & Path-VQA & 19,654 & 6,719 \\
\midrule
\multirow{1}{*}{\parbox{3cm}{Report Generation}} & Chest X-ray & MIMIC-CXR & 25,000 & 3,513 \\
\midrule
\multirow{2}{*}{\parbox{3cm}{Image Captioning}} & Radiology & ROCO & 25,000 & 8,786 \\
& Radiology & IU X-RAY & 6,698 & 745 \\
 & Pathology & Peir Gross & 5,172 & 1,289\\
\bottomrule
\end{tabular}
\end{table}

\begin{table}[htbp]
    \centering
    \caption{Selected hyperparameters for Stage 1 and Stage 2 training of Dragonfly-Med.}
    \label{tab:hyperparams_biomed}
    \begin{tabular}{@{}lcc@{}}
        \toprule
        Hyperparameter                & Stage 1         & Stage 2         \\
        \midrule
        Batch Size                    & 64               & 16               \\
        Learning Rate                 & 2e-5            & 2e-6            \\
        LR Scheduler                  & cosine          & cosine          \\
        Warmup Steps Ratio            & 0.01            & 0.01            \\
        Max Sequence Length           & 4096            & 4096            \\
        Tune Projection Layer         & \checkmark      & \checkmark      \\
        Tune Vision Encoder           & \checkmark      & \checkmark      \\
        Tune LLM                      & \texttimes      & \checkmark      \\
        \bottomrule
    \end{tabular}
\end{table}

\end{document}